\documentclass[runningheads]{llncs}
\usepackage{adjustbox}
\usepackage{amssymb}
\usepackage{amsmath}
\usepackage{array, boldline, makecell, booktabs}
\usepackage{epsfig}
\usepackage[mathscr]{euscript}
\usepackage{longtable}
\usepackage{lscape}
\usepackage{multirow}
\usepackage[figuresright]{rotating}
\usepackage{subfigure}
\usepackage[para,online,flushleft]{threeparttable}
\usepackage{tabularx}
\usepackage{xcolor}
\usepackage{url}
\usepackage[boxed,ruled,commentsnumbered]{algorithm2e}

\makeatletter
\newcommand{\printfnsymbol}[1]{%
  \textsuperscript{\@fnsymbol{#1}}%
}
\makeatother

\begin{document}

\title{$\mathsf{SAFE}$: Similarity-Aware Multi-Modal Fake News Detection \vspace{-0.5em}}

\author{Xinyi Zhou\thanks{Authors are equally contributed.}
%\inst{1}
%\orcidID{0000-0002-2388-254X} 
\and
Jindi Wu\printfnsymbol{1} \and
Reza Zafarani 
%\inst{1}
\vspace{-0.5em}}
\authorrunning{X. Zhou et al.}

\institute{Data Lab, EECS Department, Syracuse University, NY 13244, U.S.A. \\
\email{\{zhouxinyi,reza\}@data.syr.edu},  \email{jwu172@syr.edu} \\
% \url{http://www.springer.com/gp/computer-science/lncs} 
}

\maketitle              % typeset the header of the contribution

\vspace{-1.5em}
\begin{abstract}
% The abstract should briefly summarize the contents of the paper in 150--250 words.

Effective detection of fake news has recently attracted significant attention. Current studies have made significant contributions to predicting fake news with less focus on exploiting the relationship (similarity) between the textual and visual information in news articles. Attaching importance to such similarity helps identify fake news stories that, for example, attempt to use irrelevant images to attract readers' attention. In this work, we propose a $\mathsf{S}$imilarity-$\mathsf{A}$ware $\mathsf{F}$ak$\mathsf{E}$ news detection method ($\mathsf{SAFE}$) which investigates multi-modal (textual and visual) information of news articles. First, neural networks are adopted to separately extract textual and visual features for news representation. We further investigate the relationship between the extracted features across modalities. Such representations of news textual and visual information along with their relationship are jointly learned and used to predict fake news. The proposed method facilitates recognizing the falsity of news articles based on their text, images, or their ``mismatches.'' We conduct extensive experiments on large-scale real-world data, which demonstrate the effectiveness of the proposed method.

\vspace{-1em}
\end{abstract}
\vspace{-2em}
\section{Introduction}

\vspace{-0.8em}
Following the 2016 U.S. presidential election, the impact of ``fake news'' has become a major concern. Based on a broad investigation of $\sim$126,000 verified true and fake news stories on Twitter from 2006 to 2017, Vosoughi and colleagues revealed that fake news stories spread more frequently and faster compared to true news stories~\cite{vosoughi2018spread}. As indicated by the fundamental theories on fake news in psychology and social sciences (see a comprehensive survey in Ref.~\cite{zhou2018survey}), the more a fake news article spreads, the higher the possibility of social media users spreading and trusting it due to repeated exposure.
%(i.e., \textit{validity effect}~\cite{hasher1977frequency}) 
and/or peer pressure. 
%(i.e., \textit{bandwagon effect}~\cite{asch1955opinions}). 
Such levels of trust and beliefs can easily be amplified and reinforced within social media due to its \textit{echo chamber effect}~\cite{jamieson2008echo}.
% When a fake news article has completely gained the trust of a social media user, revising or correcting such beliefs is often difficult, even when users are presented with new evidence. 
% (due to \textit{conservatism bias}~\cite{basu1997conservatism}).
%and \textit{Semmelweis reflex}~\cite{balint2009semmelweis}). 
Hence, extensive research has been conducted on effective detection of fake news to block its dissemination on social media. Fake news detection methods can be generally grouped into (1) content-based and (2) social-context-based methods.
% lazer2018science
The main difference between the two types of methods is whether or not they rely on \textit{social context} information: the information on how the news has propagated on social media, where abundant auxiliary information of social media users involved and their connections/networks can be utilized. Many innovative and significant solutions (e.g., \cite{castillo2011information,ruchansky2017csi,qian2018neural}) have been proposed to exploit social context information. With more social context information available, one can often better detect fake news; however, detection becomes more challenging depending on the stage the news is currently at. It is difficult to detect fake news using social-context-based methods when it has been just published and has not been propagated (i.e., no social context information), which motivates us to further explore the role that news content can play in fake news detection.

As ``a news article that is intentionally and verifiably false''~\cite{reza2019tutorial}, fake news content often contains textual and visual information. Existing content-based fake news detection methods either solely consider textual information~\cite{zhou2019content}, or combine both types of data ignoring the relationship (similarity) between them~\cite{wang2018eann,yang2018ti,jin2017novel,jin2017multimodal}. The values in understanding such relationship (similarity) for predicting fake news are two-fold. To attract public attention, some fake news stories (or news stories with low-credibility) prefer to use dramatic, humorous (facetious), and tempting images whose content is far from the actual content within the news text. 
% {\color{red}An example of such type of fake news can be seen in Fig. \ref{fig:fn1}}. % Compared to text, in particular, long articles, readers' attention is often first given to images that can reveal the topics 
Furthermore, when a fake news article tells a story with fictional scenarios or statements, it is difficult to find both pertinent and non-manipulated images to match these fictions; hence a ``gap'' exists between the textual and visual information of fake news when creators use non-manipulated images to support non-factual scenarios or statements.\footnote{Examples at \url{https://www.snopes.com/fact-check/rating/miscaptioned/}.}

With such considerations, we propose a $\mathsf{S}$imilarity-$\mathsf{A}$ware $\mathsf{F}$ak$\mathsf{E}$ news detection method ($\mathsf{SAFE}$). The method consists of three modules, performing (1) multi-modal (textual and visual) feature extraction; (2) within-modal (or say, modal-independent) fake news prediction; (3) cross-modal similarity extraction, respectively. For each news article, we first adopt neural networks to automatically obtain the latent representation of both its textual and visual information, based on which a similarity measure is defined between them. Then, such representations of news textual and visual information with their similarity are jointly learned and used to predict fake news. The proposed method aims to recognize the falsity of a news article on either its text or images, or the ``mismatch'' between the text and images.

The main contributions of our work are summarized as below.

\vspace{-0.7em}
\begin{enumerate}
    \item To our best knowledge, we present the first approach that investigates the role of the relationship (similarity) between news textual and visual information in predicting fake news; 
    % We investigate multi-modal (textual and visual) information of news articles, as well as the role of their relationship in predicting fake news.
    \item We propose a new method to jointly exploit multi-modal (textual and visual) and relational information to learn the representation of news articles and predict fake news; and
    \item We conduct extensive experiments on large-scale real-world data to demonstrate the effectiveness of the proposed method.
\end{enumerate}

\vspace{-0.7em}
Next, we will first review the related work in Sec. \ref{sec:related_work}. The proposed method will be detailed in Sec. \ref{sec:method}, along with its iterative learning process in Sec. \ref{sec:optimization}. We will detail the experiments and the results in Sec. \ref{sec:experiment}. We will conclude in Sec. \ref{sec:conclusion}.

\section{Related Work}
\label{sec:related_work}
% Fake news can be detected before it has been propagated, e.g., when it is published on a news outlet such as CNN

\vspace{-0.8em}
There has been extensive research on fake news detection. Fake news detection methods can be generally grouped into (I) content-based and (II) social-context-based methods.

\vspace{-1.5em}
\subsubsection{I. Content-based Fake News Detection}
Content-based methods detect fake news by utilizing news content, i.e., the \textit{textual information} and/or \textit{visual information} within news content. 

Most content-based methods have comprehensively investigated news textual information. Within a traditional statistical natural language processing framework, such investigation has crossed multiple levels of language. By assuming that fake news differs from true news in linguistic/writing styles in the content, various hand-crafted features have been extracted from news content for representation and used for classification by, e.g., SVM and random forest.
% and XGBoost~\cite{chen2016xgboost}. 
For example, P\'erez-Rosas et al. employed lexical features by using bag-of-words with $n$-gram models, semantic features relying on LIWC~\cite{pennebaker2015development}, syntactic features such as context-free grammars, and news readability~\cite{perez2017automatic}. Instead of extracting features based on experience, Zhou et al.~\cite{zhou2019content} validated the role of fundamental theories in psychology and social science in guiding fake news feature engineering. 
% attributes such as the sentiment, quality, and quantity within news content were thus investigated, and some linguistic patterns were revealed in such way~\cite{zhou2019content}.
Rhetorical structures among sentences or phrases within news content have also been investigated with either a vector space model~\cite{rubin2015truth} or Bi-LSTM~\cite{karimi2019learning}. Researchers have also explored the political bias~\cite{potthast2017stylometric} and homogeneity~\cite{horne2019different} of news publishers by mining news content that they have published, and have demonstrated how such information can help detect fake news.
% the former proposed a new way of assessing linguistic style similarity between two pieces of news content via Unmasking - a meta-learning approach; and the latter discovered that content sharing happens in tightly formed communities and these communities represent relatively homogeneous portions of the media landscape, where effects such as echo chambers have been validated.
% Zhang et al. constructed a tripartite graph among news authors, content and topics~\cite{zhangfakedetector}, and conducted news verification by the proposed gated diffusive units.

In addition to textual information, greater -- while still limited -- attention has been recently paid to visual information within news content. Jin et al. analyzed images between true news and fake news in terms of, e.g., their clarity~\cite{jin2017novel}. Along with the recent advances in deep learning, various RNNs and CNNs have been developed for multi-modal fake news detection and related tasks~\cite{jin2017multimodal,wang2018eann,khattar2019mvae,truong2019multimodal,wang2019learning,yang2018ti}. To learn the multi-modal (textual and visual) representation of news content, Jin et al. developed VGG-19 and LSTM with an attention mechanism~\cite{jin2017multimodal}, and Khattar et al. designed an encoder-decoder mechanism~\cite{khattar2019mvae}.
% For example, Jin et al. adopted VGG-19 and LSTM with attention mechanism to learn the multi-modal (textual and visual) representation of news content in a concatenation way~\cite{jin2017multimodal}.
Yang et al. proposed TI-CNN, which detects fake news by extracting both explicit and latent multi-modal features within news content~\cite{yang2018ti}.
Wang et al. proposed Event Adversarial Neural Network (EANN) to learn event-invariant features representative of news content across various topics and domains~\cite{wang2018eann}. 
% Khattar et al. applied an encoder-decoder mechanism to learn the representation of multi-modal information within news content~\cite{khattar2019mvae}. 
While current techniques have facilitated the development of multi-modal fake news detection, the relationship across modalities has been barely explored and exploited.  
Our work bridges this gap by directly capturing the relationship (similarity) between the textual and visual information within news content, and firstly learning the representation of news articles through mining its multi-modal information and the relationship across modalities.

% wu2015false,ma2018rumor\
\vspace{-1.5em}
\subsubsection{II. Social-context-based Fake News Detection} Social-context-based methods detect fake news by investigating social-context information related to news articles, i.e., how news articles spread on social media. Significant contributions have been made on identifying the differences in propagation patterns between fake news and the truth~\cite{vosoughi2018spread}. Such contributions have also focused on how user profiles~\cite{castillo2011information} and opinions~\cite{qian2018neural,ruchansky2017csi} can help news verification using feature engineering~\cite{castillo2011information} and neural networks~\cite{qian2018neural,ruchansky2017csi}. Nevertheless, verifying a news article that has been published online, e.g., on a news outlet such as BuzzFeed (\url{buzzfeed.com}), before it has been disseminated on social media demands content-based methods as social-context information at this stage does not exist. For this purpose, we focus on mining news content in this work, where the proposed method will be detailed next.

\begin{figure}[t]
    \centering
    \includegraphics[width=\textwidth]{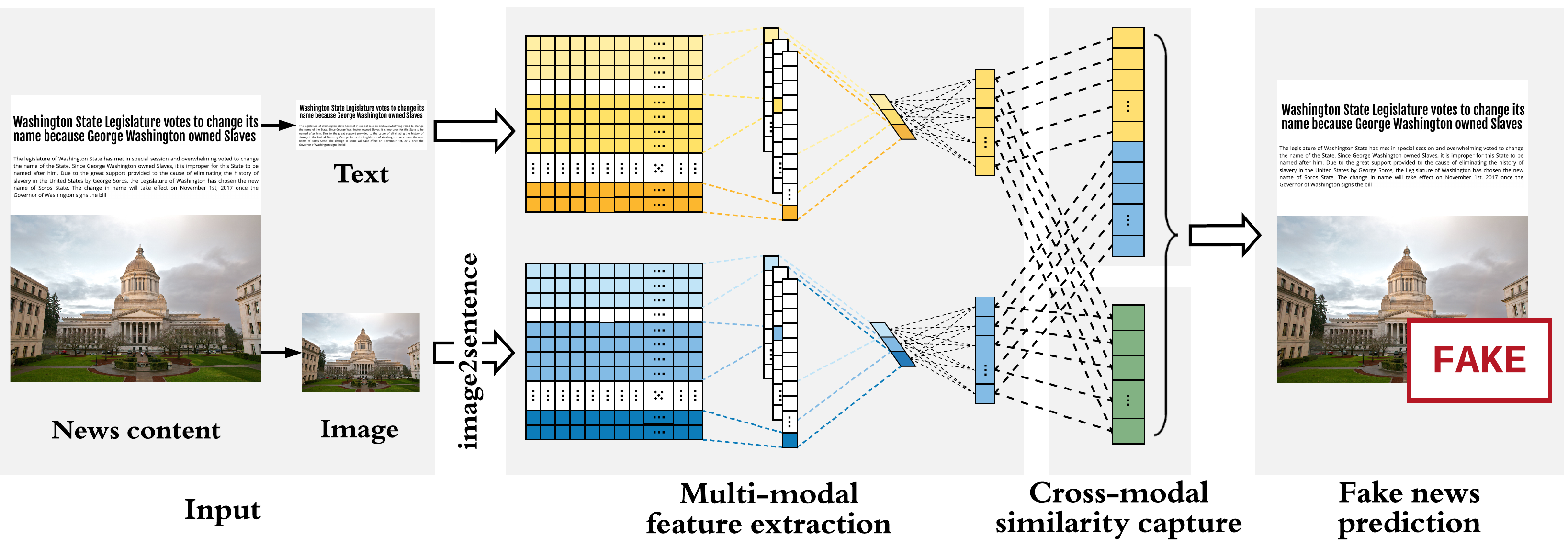}\vspace{-1em}
    \caption{Overview of the $\mathsf{SAFE}$ framework}
    \vspace{-1.5em}
    \label{fig:framework}
\end{figure}

\vspace{-1.5em}
\section{Methodology}
\label{sec:method}

\vspace{-1em}
In this section, the proposed method ($\mathsf{SAFE}$) is detailed in terms of its three modules performing: (I) multi-modal feature extraction (Sec. \ref{subsec:feature_extraction}), (II) modal-independent fake news prediction (Sec. \ref{subsec:within_model}), and (III) cross-modal similarity extraction (Sec. \ref{subsec:corss_modal}). Then, we detail in Sec. \ref{subsec:integration} how various modules can work collectively to predict fake news. An overview of the $\mathsf{SAFE}$ framework is presented in Fig. \ref{fig:framework}. Before further specification, we formally define the problem and introduce some key notations as follows. 

\vspace{-1.5em}
\subsubsection{Problem Definition and Key Notation.}
Given a news article $A = \{T, V\}$ consisting of textual information $T$ and visual information $V$, we denote $\mathbf{t} \in \mathbb{R}^{d}$ and $\mathbf{v} \in \mathbb{R}^{d}$ as the corresponding representations, where $\mathbf{t} = \mathcal{M}_t(T,\theta_t)$ and $\mathbf{v} = \mathcal{M}_v(V,\theta_v)$. Let $s =  \mathcal{M}_s(\mathbf{t},\mathbf{v})$ denote the similarity between $\mathbf{t}$ and $\mathbf{v}$, where $s \in [0,1]$. Our goal is to predict whether $A$ is a fake news article ($\hat{y}=1$) or a true one ($\hat{y}=0$) by investigating its textual information, visual information, and their relationship, i.e., to determine $\mathcal{M}_p: (\mathcal{M}_t,\mathcal{M}_v,\mathcal{M}_s) \xrightarrow{(\theta_t, \theta_v, \theta_p)} \hat{y} \in \{0,1\}$, where $\theta_*$ are parameters to be learned.\vspace{-1em}
% i.e., $\{T, V\} \overset{\Theta}{\rightarrow} y \in \{0,1\}$.

\vspace{-0.5em}
\subsection{Multi-modal Feature Extraction}
\label{subsec:feature_extraction}

\vspace{-0.5em}
The multi-modal feature extraction module of $\mathsf{SAFE}$ aims to represent the (I) textual information and (II) visual information of a given news article in $d$-dimensional space, respectively. 

\vspace{-1.5em}
\subsubsection{Text} 
\begin{figure}[t]
    \centering
    \includegraphics[width=.88\textwidth]{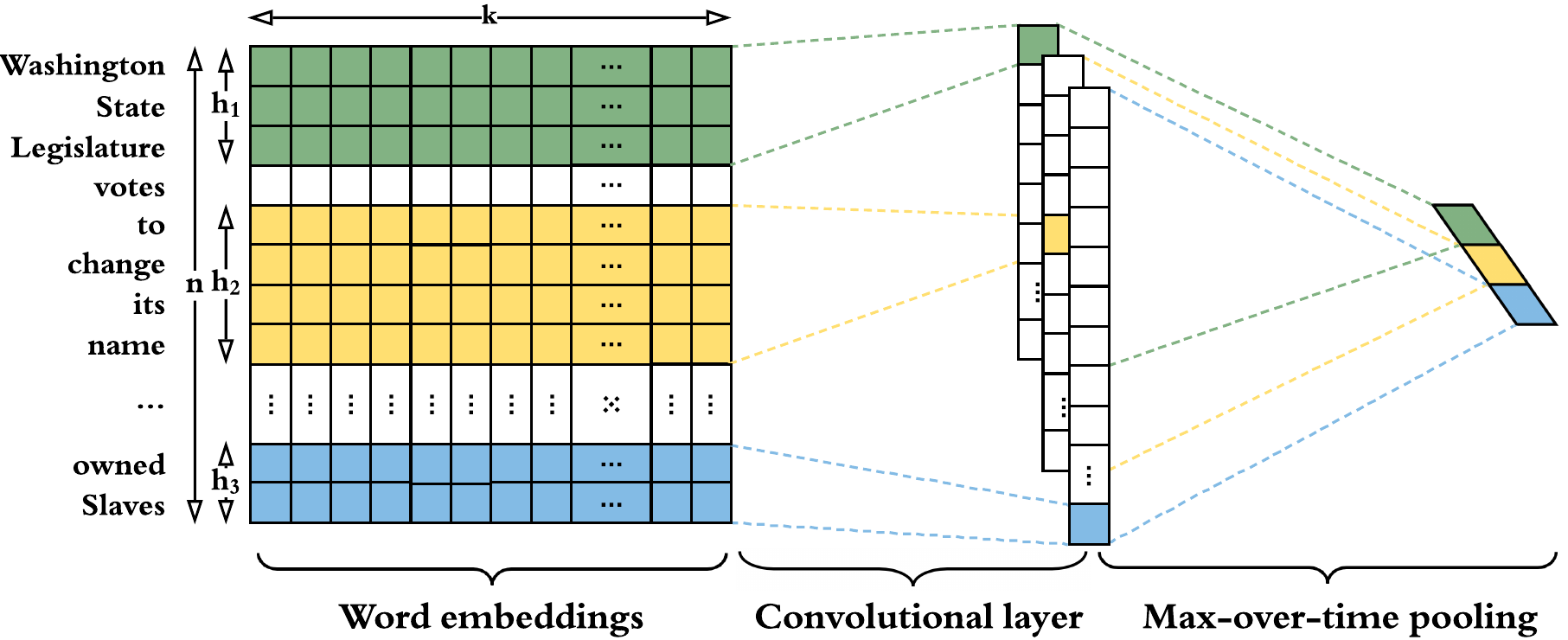}\vspace{-1em}
    \caption{Text-CNN Architecture}
    \vspace{-1.25em}
    \label{fig:textcnn}
\end{figure}

We extend Text-CNN~\cite{kim2014convolutional} by introducing an additional fully connected layer to automatically extract textual features for each news article. The architecture of Text-CNN is provided in Fig. \ref{fig:textcnn}, which contains a convolutional layer and max pooling. 
Given a piece of content with $n$ words, each word is first embedded as $\mathbf{x}_t^l \in \mathbb{R}^{k}, l=1,2,\cdots,n$~\cite{mikolov2013efficient}. The convolutional layer is used to produce a \textit{feature map}, denoted as $C_t = \{c_t^i\}_{i=1}^{n-h+1}$, from a sequence of local inputs $\{\mathbf{x}_t^{i:(i+h-1)}\}_{i=1}^{n-h+1}$, % $i=1,2,\cdots,n-h+1$ 
via a \textit{filter} $\mathbf{w}_t$. As shown in Fig. \ref{fig:textcnn}, each local input is a group of $h$ continuous words. Mathematically,
\vspace{-0.8em}
\begin{equation}
\vspace{-0.5em}
\label{eq:filter}
c_t^i = \sigma (\mathbf{w}_t \cdot  \mathbf{x}_t^{i:(i+h-1)}+b_t),
\end{equation}
\vspace{-1.5em}
\begin{equation}
\vspace{-0.5em}
\mathbf{x}_{i:(i+h-1)} = \mathbf{x}_i \oplus \mathbf{x}_{i+1} \oplus \cdots \oplus \mathbf{x}_{i+h-1},
\end{equation}
where $\mathbf{w}_t, \mathbf{x}_t^{i:(i+h-1)} \in \mathbb{R}^{hk}$, $b_t \in \mathbb{R}$ is a \textit{bias},
$\oplus$ is the concatenation operator, and
$\sigma$ is ReLU
% ~\cite{nair2010rectified} 
function. Note that $\mathbf{w}_t$ and $b_t$ are all parameters within Text-CNN to be learned. Then, a max-over-time pooling operation is applied on the obtained feature map for dimension reduction, i.e., $\hat{c}_t = \max\{c_t^i\}_{i=1}^{n-h+1}$.
Finally, the representation of the news text can be obtained by $\mathbf{t} = \mathbf{W}_t \mathbf{\hat{c}}_t+\mathbf{b}_t$, where $\mathbf{\hat{c}}_t \in \mathbb{R}^{g}$, $g$ is the different number of window sizes chosen; $\mathbf{W}_t \in \mathbb{R}^{d \times g}$ and $\mathbf{b}_t \in \mathbb{R}^{d}$ are parameters to be learned.

\vspace{-1.5em}
\subsubsection{Image} For representing news images, we also use Text-CNN with an additional fully connected layer while we first process visual information within news content using a pre-trained \texttt{image2sentence} model\footnote{\url{https://github.com/nikhilmaram/Show_and_Tell}}~\cite{vinyals2016show}. Compared to existing multi-modal fake news detection studies that often directly apply a pre-trained CNN (e.g., VGG) model to obtain the representation of news images~\cite{wang2018eann,jin2017multimodal}, we adopt the aforementioned processing strategy for consistency and to increase insights when computing the similarity across modalities. As we will demonstrate later in our experiments, it also leads to performance improvements.
%(i.e., entries in both $\mathbf{t}$ and $\mathbf{v}$ sequentially capture the common flow of ), in particular, when assessing the similarity between news textual and visual information (see Section \ref{subsec:corss_modal})\footnote{Such consistency does help }.  
% For the representation of news images, CNN is used with a particular choice of VGG-19 architecture as well as an additional full connected layer for dimension transformation. 
% See Fig. \ref{subfig:vgg19} for the architecture of VGG-19. Given the pixel RGB tensor of a news image as the input of the neural network, 
Let $\mathbf{\hat{c}}_v$ denote the output of the neural network with parameters $\mathbf{w}_v$ (filter) and $b_v$ (bias). Similarly, the final representation of news visual information is then computed by  $\mathbf{v} = \mathbf{W}_v \mathbf{\hat{c}}_v+\mathbf{b}_v$, where $\mathbf{W}_v$ and $\mathbf{b}_v$ are parameters to be learned.

\vspace{-1.5em}
\subsection{Modal-independent Fake News Prediction}
\label{subsec:within_model}

\vspace{-0.5em}
To properly represent news textual and visual information in predicting fake news, we aim to correctly map the extracted textual and visual features of news content to their possibilities of being fake, and further to their actual labels. Mathematically, such possibilities can be computed by
\vspace{-0.8em}
\begin{equation}
\label{eq:softmax}
    \mathcal{M}_p(\mathbf{t},\mathbf{v}) = \mathbf{1} \cdot \text{softmax}(\mathbf{W}_p (\mathbf{t} \oplus \mathbf{v}) + \mathbf{b}_p),
    \vspace{-0.5em}
\end{equation}
where $\mathbf{1} = [1, 0]^\top$, $\oplus$ is the concatenation operator, $\mathbf{W}_p \in \mathbb{R}^{2\times 2d}$ and $\mathbf{b}_p \in \mathbb{R}^2$ are parameters. To let the computed possibilities of news articles being fake approach their actual labels, a cross-entropy-based loss function is defined:
\vspace{-0.5em}
\begin{equation}
    \mathcal{L}_p(\theta_t, \theta_v, \theta_p) = -\mathbb{E}_{(a,y)\sim (A, Y)} (y \log \mathcal{M}_p(\mathbf{t}, \mathbf{v}) + (1-y) \log (1- \mathcal{M}_p(\mathbf{t}, \mathbf{v}))),
    \vspace{-0.5em}
\end{equation}
where $\theta_p = \{ \mathbf{W}_p, \mathbf{b}_p \}$, 
$\theta_t = \{ \mathbf{W}_t, \mathbf{b}_t, \mathbf{w}_t, b_t \}$, $\theta_v = \{ \mathbf{W}_v, \mathbf{b}_v, \mathbf{w}_v, b_v \}$, and
\vspace{-0.7em}
\begin{equation}
    (\hat{\theta}_t, \hat{\theta}_v, \hat{\theta}_p) = \arg\underset{\theta_t, \theta_v, \theta_p}{\min}~\mathcal{L}_p(\theta_t,\theta_v,\theta_p).
\end{equation}

\vspace{-2.2em}
\subsection{Cross-modal Similarity Extraction}
\label{subsec:corss_modal}

\vspace{-0.5em}
When attempting to correctly map the multi-modal features of news articles to their labels, features belonging to two different modals are considered separately -- concatenating them with no relation between them explored (see Sec. \ref{subsec:within_model}). However, besides that, the falsity of a news article can be also detected by assessing how (ir)relevant the textual information is compared to its visual information; fake news creators sometimes actively use irrelevant images for false statements to attract readers' attention, or passively use them due to the difficulty in finding a supportive non-manipulated image (see case studies in Sec. \ref{sec:experiment} for examples).
Compared to news articles delivering relevant textual and visual information, those with disparate statements and images are more likely to be fake. We define the relevance between news textual and visual information as follows by slightly modifying cosine similarity:
\vspace{-0.7em}
\begin{equation}
    \mathcal{M}_s(\mathbf{t}, \mathbf{v}) = 
    \frac{\mathbf{t} \cdot \mathbf{v} + \left \| \mathbf{t} \right \| \left \| \mathbf{v} \right \|}{2\left \| \mathbf{t} \right \| \left \| \mathbf{v} \right \|}
    % \frac{\mathbf{t} \cdot \mathbf{v}}{2\left \| \mathbf{t} \right \| \left \| \mathbf{v} \right \|}+\frac{1}{2}
    % \sqrt{\mathbf{t} \cdot \mathbf{t}} \sqrt{\mathbf{v} \cdot \mathbf{v}}}{2\sqrt{\mathbf{t} \cdot \mathbf{t}} \sqrt{\mathbf{v} \cdot \mathbf{v}}}
    \vspace{-0.7em}
\end{equation}
% where $\cos(\mathbf{t}, \mathbf{v})$ denotes the cosine similarity between $\mathbf{t}$ (textual features) and $\mathbf{v}$ (visual features) computed by:
% \vspace{-0.5em}
% \begin{equation}
%          \cos(\mathbf{t}, \mathbf{v})= \frac{\mathbf{t} \cdot \mathbf{v}}{\sqrt{\mathbf{t}^\top \mathbf{t}} \sqrt{\mathbf{v}^\top \mathbf{v}}}
%          % \cos(\mathbf{t}, \mathbf{v})=  \frac{\sum_{i=1}^{d} \mathbf{t}_i \mathbf{v}_i}{\sqrt{\sum_{i=1}^{d} \mathbf{t}_i^2} \sqrt{\sum_{i=1}^{d} \mathbf{v}_i^2}},
%          \vspace{-0.5em}
% \end{equation}
In such a way, it is guaranteed that $\mathcal{M}_s(\mathbf{t}, \mathbf{v})$ is positive and $\in [0,1]$ (to be utilized in Eq. (\ref{eq:similarity_loss})); 0 indicates that $\mathbf{t}$ and $\mathbf{v}$ are far from being similar, while 1 indicates that $\mathbf{t}$ and $\mathbf{v}$ are exactly the same. 

Then, we can define the loss function based on cross-entropy as below, which assumes that news articles formed with mismatched textual and visual information are more likely to be fake compared to those with matching textual statements and images, when analyzing from a pure similarity perspective:
\vspace{-0.5em}
\begin{equation}
\label{eq:similarity_loss}
    \mathcal{L}_s(\theta_t, \theta_v) = -\mathbb{E}_{(a,y)\sim (A, Y)} (y \log (1-\mathcal{M}_s(\mathbf{t}, \mathbf{v})) + (1-y) \log \mathcal{M}_s(\mathbf{t}, \mathbf{v})),
\end{equation}
\vspace{-1.5em}
\begin{equation}
    (\hat{\theta}_t, \hat{\theta}_v) = \arg\underset{\theta_t, \theta_v}{\min}~\mathcal{L}_s(\theta_t,\theta_v).
\end{equation}

\vspace{-1.8em}
\subsection{Model Integration and Joint Learning}
\label{subsec:integration}

\vspace{-0.3em}
When detecting fake news, we aim to correctly recognize fake news stories whose falsity is in their (1) textual and/or visual information, or (2) their relationship, as specified in Sec. \ref{subsec:within_model} and Sec. \ref{subsec:corss_modal}, respectively. To involve both cases, we specify our final loss function as
\vspace{-0.5em}
\begin{equation}
\label{eq:loss_function}
    \mathcal{L}(\theta_t,\theta_v,\theta_p) = \alpha \mathcal{L}_p(\theta_t,\theta_v,\theta_p) + \beta \mathcal{L}_s(\theta_t,\theta_v),
    \vspace{-0.5em}
\end{equation}
where parameters can be jointly learned by
\vspace{-0.5em}
\begin{equation}
\label{eq:object}
    (\hat{\theta}_t, \hat{\theta}_v, \hat{\theta}_p) = \arg\underset{\theta_t, \theta_v, \theta_p}{\min}~\mathcal{L}(\theta_t,\theta_v,\theta_p).
    \vspace{-0.5em}
\end{equation}

% Finally, given a news article, its label $\hat{y}$ can be predicted by the softmax classifier in the proposed method with all learned parameters.

%\begin{minipage}{0.55\textwidth}
%\begin{algorithm}[H]
\begin{algorithm}[t]
\small
\LinesNumbered
\caption{$\mathsf{SAFE}$}
\label{alg:safe}
\KwIn{$A = \{ (T_j, V_j)\}_{j=1}^m$, $Y = \{ y_j \}_{j=1}^m$, $H = \{ h_k \}_{k=1}^g$, $\gamma$}
% \KwOut{$\theta_p$, $\theta_t$, $\theta_v$}
\KwOut{$\theta_p = \{ \mathbf{W}_p, \mathbf{b}_p \}$, $\theta_t = \{\mathbf{W}_t, \mathbf{b}_t, \mathbf{w}_t, b_t \}$, $\theta_v = \{\mathbf{W}_v, \mathbf{b}_v, \mathbf{w}_v, b_v \}$}
% Randomly initialize $ \mathbf{W}_p, \mathbf{W}_t, \mathbf{W}_v, \mathbf{b}_p, \mathbf{b}_t, \mathbf{b}_v, \mathbf{w}_t, \mathbf{w}_v,  b_t, b_v $ \;
% \ForEach{$(T_j, V_j)$}{
%       $\mathbf{x}^{1:n}_t \leftarrow \texttt{word2vec}(T_j)$\;
%       $\mathbf{x}^{1:n}_v \leftarrow \texttt{word2vec}(\texttt{image2sentence}(V_j))$\;
%       }
Randomly initialize $\mathbf{W}_p, \mathbf{b}_p, \mathbf{W}_t, \mathbf{b}_t, \mathbf{w}_t, b_t, \mathbf{W}_v, \mathbf{b}_v, \mathbf{w}_v, b_v$\;
\While{not convergence}{
    \ForEach{$(T_j, V_j)$}{
    
      Update $\theta_p$: $\{ \mathbf{W}_p, \mathbf{b}_p \} \leftarrow$ Eq. (\ref{eq:update_theta_p2})\;
    %   \qquad $\mathbf{b}_p \leftarrow \mathbf{b}_p - \gamma \cdot \alpha (\mathcal{M}_p - y_j)$ \;
      
      \ForEach{$h_k$}{
      Update $\theta_{t}$: $\{ \mathbf{W}_{t}, \mathbf{b}_{t}, \mathbf{w}_{t}, b_{t} \} \leftarrow$ Eqs. (\ref{eq:lp2t}-\ref{eq:bt})\;
      % \qquad $\mathbf{w}_{t}, b_{t} \leftarrow$ Eq. (\ref{eq:update_theta_t2})

      Update $\theta_v$: similar to updating $\theta_t$\;
      }
    }
}
\Return{$\mathbf{W}_p, \mathbf{b}_p, \mathbf{W}_t, \mathbf{b}_t, \mathbf{w}_t, b_t, \mathbf{W}_v, \mathbf{b}_v, \mathbf{w}_v, b_v$} 
\end{algorithm} 
% \end{minipage}

\vspace{-2em}
\section{Optimization}
\label{sec:optimization}

\vspace{-0.5em}
We outline the optimization process to learn the model parameters, i.e., iteratively solving Eq. (\ref{eq:object}). The process is summarized in Algorithm \ref{alg:safe}. The updating rule for each parameter is as follows:

\vspace{-1em}
\paragraph{\emph{\textbf{Update $\theta_{p}$.}}} Let $\gamma$ be the learning rate, the partial derivative of $\mathcal{L}$ w.r.t. $\theta_p$ is:
\vspace{-0.8em}
\begin{equation}
\label{eq:update_theta_p}
    \theta_p \leftarrow \theta_p - \gamma \cdot \alpha \frac{\partial \mathcal{L}_p}{\partial \theta_p}.
    \vspace{-0.8em}
\end{equation}

As $\theta_p = \{ \mathbf{W}_p, \mathbf{b}_p \}$, updating $\theta_p$ is equivalent to updating both $\mathbf{W}_p$ and $\mathbf{b}_p$ in each iteration, which respectively follow the following rules:
\vspace{-0.7em}
\begin{equation}
    \label{eq:update_theta_p2}
    \begin{matrix}
    \mathbf{W}_p \leftarrow \mathbf{W}_p - \gamma \cdot \alpha  \Delta \mathbf{y} (\mathbf{t} \oplus \mathbf{v})^\top,\quad
    \mathbf{b}_p \leftarrow \mathbf{b}_p - \gamma \cdot \alpha  \Delta \mathbf{y},
    \end{matrix}
    \vspace{-0.8em}
\end{equation}
where $\Delta \mathbf{y} = \left [ \hat{y} - y, y - \hat{y} \right]^\top$.

\vspace{-0.8em}
\paragraph{\emph{\textbf{Update $\theta_{t}$.}}} The partial derivative of $\mathcal{L}$ w.r.t. $\theta_t$ is generally computed by
% To update $\theta_t = \{ \mathbf{W}_t, \mathbf{b}_t, \mathbf{w}_t, b_t \}$ requests
\vspace{-0.6em}
\begin{equation}
    \label{eq:update_theta_t}
    % \begin{array}{lll}
    % \theta_{t} & \leftarrow & \theta_{t} - \gamma (\alpha \frac{\partial \mathcal{L}_p}{\partial \theta_{t}} + \beta \frac{\partial \mathcal{L}_s}{\partial \theta_{t}}) \\
    \theta_{t} \leftarrow \theta_{t} - \gamma (\alpha \frac{\partial \mathcal{L}_p}{\partial \mathcal{M}_t} \frac{\partial \mathcal{M}_t}{\partial \theta_{t}} + \beta \frac{\partial \mathcal{L}_s}{\partial \mathcal{M}_t} \frac{\partial \mathcal{M}_t}{\partial \theta_{t}}).
    % \end{array}
    \vspace{-0.7em}
\end{equation}
Let $\nabla \mathcal{L}_*(\mathbf{t}) =  \frac{\partial \mathcal{L}_*}{\partial \mathcal{M}_t}$, $\mathbf{t}_0 = \frac{\mathbf{t}}{|\left | \mathbf{t} |\right | }$, $\mathbf{v}_0 = \frac{\mathbf{v}}{|\left | \mathbf{v} |\right | }$, and $\mathbf{W}_{p,L}$ denote the first $d$ columns of $\mathbf{W}_{p}$, we can have
\vspace{-0.6em}
\begin{equation}
\label{eq:lp2t}
    \nabla \mathcal{L}_p(\mathbf{t}) = \mathbf{W}_{p,L}^\top \Delta \mathbf{y},
\end{equation}
\vspace{-1.3em}
\begin{equation}
\nabla \mathcal{L}_s(\mathbf{t})  = \frac{1 - y}{2 s \left \| \mathbf{t} \right \|} ((2s-1) \mathbf{t}_0 - \mathbf{v}_0), \vspace{-0.5em}
\end{equation}
based on which the parameters in $\theta_t$ are respectively updated as follows:
\vspace{-0.8em}
\begin{equation}
\label{eq:update_theta_t1}
      \begin{matrix}
      \mathbf{W}_{t} \leftarrow \mathbf{W}_{t} - \gamma \cdot \mathbf{D}_t \mathbf{B}_t,\quad
      \mathbf{b}_{t} \leftarrow \mathbf{b}_t - \gamma \cdot \mathbf{B}_t,
      \end{matrix} 
\end{equation}
\vspace{-1.8em}
\begin{equation}
\label{eq:update_theta_t2}
      \begin{matrix}
      \mathbf{w}_{t} \leftarrow \mathbf{w}_{t} - \gamma \cdot \mathbf{x}_{t}^{\hat{i}:(\hat{i}+h-1)} \mathbf{W}_{t}^\top \mathbf{B}_t,\quad
      b_{t} \leftarrow b_{t} - \gamma \cdot \mathbf{W}_{t}^\top \mathbf{B}_t,
      \end{matrix} 
      \vspace{-0.3em}
\end{equation}
where $\hat{i} = \arg \underset{i}{\max}\{ c_t^{i} \}_{i=1}^{n-h+1}$, $\mathbf{D}_t \in \mathbb{R}^{d \times d}$ is a diagonal matrix with entry value $c_t^{\hat{i}}$, and
\vspace{-1.3em}
\begin{equation}
\label{eq:bt}
    \mathbf{B}_t = \alpha \nabla \mathcal{L}_p(\mathbf{t}) + \beta \nabla \mathcal{L}_s(\mathbf{t}). 
\end{equation}

\vspace{-0.8em}
\paragraph{\emph{\textbf{Update $\theta_{v}$.}}} It is similar to updating $\theta_t$; we omit details due to space constraints. 

\vspace{-1em}
\section{Experiments}
\label{sec:experiment}

\vspace{-0.5em}
We detail experimental setup in Sec \ref{subsec:experimental_setup}, followed by evaluating $\mathsf{SAFE}$ in Sec \ref{subsec:performance_evaluation}.

\vspace{-1em}
\subsection{Experimental Setup}
\label{subsec:experimental_setup}

\vspace{-0.3em}
We detail (I) the data used in our experiments, (II) the baselines $\mathsf{SAFE}$ is compared to, and (III) implementation details such as how data was pre-processed and $\mathsf{SAFE}$ hyper-parameters were set.

\vspace{-1em}
\subsubsection{Datasets}
Our experiments are conducted on two well-established public benchmark datasets of fake news detection\footnote{\url{https://github.com/KaiDMML/FakeNewsNet}}~\cite{shu2018fakenewsnet}. News articles in datasets are respectively collected from PolitiFact and GossipCop. 
PolitiFact (\url{politifact.com}) is a well-known non-profit fact-checking website of political statements and reports in the U.S.~\cite{wang2017liar}.
%based on which other datasets such as LIAR have been constructed as the foundation of fake news research.
GossipCop (\url{gossipcop.com}) is a website that fact-checks celebrity reports and entertainment stories published in magazines and newspapers.
News articles in PolitiFact dataset were published from May 2002 to July 2018 and those in GossipCop dataset were published from July 2000 to December 2018.
Ground truth labels (\textit{fake} or \textit{true}) of news articles in both datasets were provided by domain experts, which guarantees the quality of news labels. 
% In addition to news content and labels, both datasets also provide massive information on social network of users involved in spreading true/fake news on Twitter with their posts and comments, which are valuable for reproducing social-context-based baseline models.
Statistics of the two datasets are provided in Tab.~\ref{tab::datasets}.
% \begin{table}[h]
% \centering
% \caption{Data Statistics}
% \label{tab::datasets}
% \begin{tabular}{lrr}
% \toprule[1pt]
% \multicolumn{1}{l}{\textbf{Data}} & \multicolumn{1}{c}{\textbf{PolitiFact}} & \multicolumn{1}{c}{\textbf{GossipCop}} \\ \hline
%  \# News Stories & 1056 & 22,865 \\ 
%  \# True News & 624 & 16,817 \\ 
%  \# Fake News & 432 & 6,048 \\
%  \# Users & 384,913 & 739,166 \\
%  \# Tweets & 275,058 & 1,058,330 \\
%  \# Retweets & 293,438 & 530,833 \\
%  \# Replies & 125,654 & 232,923 \\
%   \bottomrule[1pt] 
% \end{tabular}
% \end{table}

\begin{table}[t]
\centering
\small
\caption{Data Statistics}\vspace{-0.75em}
\label{tab::datasets}
\begin{tabular}{|l|c|c|c|c|c|c|}
\hline & \multicolumn{3}{c|}{\textbf{PolitiFact}} & \multicolumn{3}{c|}{\textbf{GossipCop}} \\ \cline{2-7}
 & \multicolumn{1}{c|}{\textbf{Fake}} & \multicolumn{1}{c|}{\textbf{True}} &  \multicolumn{1}{c|}{\textbf{Overall}} & \multicolumn{1}{c|}{\textbf{Fake}} &  \multicolumn{1}{c|}{\textbf{True}} & \multicolumn{1}{c|}{\textbf{Overall}} \\ \hline
\textbf{\# News articles} & \multicolumn{1}{c|}{432} & \multicolumn{1}{c|}{624} & \multicolumn{1}{c|}{1,056} & \multicolumn{1}{c|}{5,323} & \multicolumn{1}{c|}{16,817} & \multicolumn{1}{c|}{22,140} \\  
\qquad \textbf{-- with textual information} & \multicolumn{1}{c|}{420} & \multicolumn{1}{c|}{528} & \multicolumn{1}{c|}{948} & \multicolumn{1}{c|}{4,947} & \multicolumn{1}{c|}{16,694} & \multicolumn{1}{c|}{21,641} \\  
\qquad \textbf{-- with visual information} & \multicolumn{1}{c|}{336} & \multicolumn{1}{c|}{447} & \multicolumn{1}{c|}{783} & \multicolumn{1}{c|}{1,650} & \multicolumn{1}{c|}{16,767} & \multicolumn{1}{c|}{18,417} \\ \hline 
\end{tabular}
\end{table}

\vspace{-1.5em}
\subsubsection{Baselines}
We compare to the following baselines, which detect fake news using (i) textual (LIWC~\cite{pennebaker2015development}), (ii) visual (VGG-19~\cite{simonyan2014very}), or (iii) multi-modal information (att-RNN~\cite{jin2017multimodal}).

% (I) content-based fake news detection and document classification methods ({\color{red}i.e., }), either relying on (i) single-modal information or (ii) multi-modal information; and (II) social-context-based fake news detection methods that further investigate how news articles are spread on social media ({\color{red}i.e., }).
\vspace{-1em}
\begin{itemize}
    \item \textbf{LIWC}~\cite{pennebaker2015development}: LIWC is a widely-accepted psycho-linguistics lexicon. Given a news story, LIWC can count the words in the text falling into one or more of over 80 linguistic, psychological, and topical categories. These numbers act as hand-crafted features used by, e.g., random forest, to predict fake news;
    
    % \item \textbf{Text-CNN~\cite{kim2014convolutional}:} As specified in Section \ref{subsec:feature_extraction}, Text-CNN\footnote{\url{https://github.com/DongjunLee/text-cnn-tensorflow}} extends traditional CNNs to classify news documents by capturing local and position-invariant textual features. 
    
    % \item \textbf{HAN}\footnotemark~\cite{yang2016hierarchical}: HAN (Hierarchical Attention Network) is proposed for news document classification. Based on Bi-GRUs, it encodes news textual information with word-level attention mechanism for each sentence and using sentence-level attention mechanism for each document;
    
    \item \textbf{VGG-19}\footnotemark~\cite{simonyan2014very}: VGG-19 is a widely-used CNN with 19 layers for image classification. We use a fine-tuned VGG-19 as one of the baselines; and
    
    % \item \textbf{EANN}\footnote{\url{https://github.com/yaqingwang/EANN-KDD18}}~\cite{wang2018eann}: EANN stands for Event Adversarial Neural Network for multi-modal fake news detection. The method was proposed to learn event-invariant features that are representative to news content across various topics and domains.
    
    \item \textbf{att-RNN}~\cite{jin2017multimodal}: att-RNN is a deep neural network model applicable for multi-modal fake news detection. It employs LSTM and VGG-19 with attention mechanism to fuse textual, visual and social-context features of news articles. We set the hyper-parameters the same as that in \cite{jin2017multimodal} and exclude the social-context features for a fair comparison.
    
    % \item \textbf{CSI}\footnote{\url{https://github.com/sungyongs/CSI-Code}}~\cite{ruchansky2017csi}: CSI detects fake news by investigating the textual information of news articles, the user responses received in news propagation on social media, and the sources promoted by users.
\end{itemize}

% \footnotetext[4]{\url{https://github.com/richliao/textClassifier}}

\footnotetext{{\url{https://github.com/tensorflow/models/tree/master/research/slim#pre-trained-models}}}

\vspace{-1em}
\noindent We also include the following variants of the proposed $\mathsf{SAFE}$ method:

\vspace{-1em}
\begin{itemize}
    \item \textbf{$\mathsf{SAFE} \setminus$T}: The proposed $\mathsf{SAFE}$ method without using textual information; 
    
    \item \textbf{$\mathsf{SAFE} \setminus$V}: The proposed $\mathsf{SAFE}$ method without using visual information; 
    
    \item \textbf{$\mathsf{SAFE} \setminus$S}: $\mathsf{SAFE}$ without capturing the relationship (similarity) between news textual and visual information. In this case, the extracted multi-modal features of each news article are fused by concatenating them; and
    
    \item \textbf{$\mathsf{SAFE} \setminus$W}: The proposed method when only the relationship between textual and visual information is assessed. In this case, the classifier is directly connected with the output of the cross-modal similarity extraction module, i.e., $\hat{y} \leftarrow \text{softmax}(\mathbf{W} [\mathcal{M}_s, 1-\mathcal{M}_s]^\top + \mathbf{b})$, where $\mathbf{W}$ and $\mathbf{b}$ are parameters.
\end{itemize}

\subsubsection{Implementation Details}
In our experiments, each dataset was separated into 80\% for training and 20\% for testing based on the publication dates of news articles, where newly published articles were treated as test data. five-fold cross-validation was used for model training. We set the learning rate as $10^{-4}$, the number of iterations as 100, and the strides ($H$) as $\{3, 4\}$. \vspace{-1em}
% $\mathsf{SAFE}$ Source codes are available at \url{https://github.com/forReproduction/SAFE} for reproduction.

\vspace{-0.5em}
\subsection{Performance Analysis}
\label{subsec:performance_evaluation}

\vspace{-0.5em}
We evaluate the general performance of $\mathsf{SAFE}$ by comparing it with (I) state-of-the-art fake news detection methods and (II) its variants. Next, (III) parameters within $\mathsf{SAFE}$ are analyzed and (IV) case studies are presented to validate its effectiveness. We use accuracy, precision, recall, and $F_1$ score to evaluate how well the representation and prediction perform.

\begin{table}[t]
\centering
% \small
\caption{Performance of Methods in Detecting Fake News}
\label{tab:general_performance}
\begin{adjustbox}{max width = \textwidth}
\begin{threeparttable}
\begin{tabular}{|l|l|c|c|c||c|c|c|c||c|}
\hline
\multicolumn{2}{|l|}{} & \textbf{LIWC}$^\dagger$  & \textbf{VGG-19}$^\wr$ & \textbf{att-RNN}$^\ddagger$ & \textbf{$\mathsf{SAFE} \setminus$T}$^\wr$ & \textbf{$\mathsf{SAFE} \setminus$V}$^\dagger$ & 	 \textbf{$\mathsf{SAFE} \setminus$S}$^\ddagger$ & \textbf{$\mathsf{SAFE} \setminus$W}$^\ddagger$ & \textbf{$\mathsf{SAFE}$}$^\ddagger$ \\ \hline
 & \textbf{Acc.} & 0.822 & 0.649 & 0.769 & 0.674 & 0.721 & 0.796 & 0.738 & \textbf{0.874} \\ \cline{2-10} 
\multirow{2}{1.3cm}{\textbf{Politi- Fact}}  & \textbf{Pre.} & 0.785 & 0.668 & 0.735 & 0.680 & 0.740 & 0.826 & 0.752 & \textbf{0.889} \\ \cline{2-10} 
 & \textbf{Rec.} & 0.846 & 0.787 & \textbf{0.942} & 0.873 & 0.831 & 0.801 & 0.844 & 0.903 \\ \cline{2-10} 
 & \textbf{$\mathbf{F}_1$} & 0.815 & 0.720 & 0.826 & 0.761 & 0.782 & 0.813 & 0.795 & \textbf{0.896} \\ \hline \hline
 & \textbf{Acc.} & 0.836 & 0.775 & 0.743 & 0.721 & 0.802 & 0.814 & 0.812 & \textbf{0.838} \\ \cline{2-10} 
\multirow{2}{1.3cm}{\textbf{Gossip- Cop}}  & \textbf{Pre.} & \textbf{0.878} & 0.775 & 0.788 & 0.734 & 0.853 & 0.875 & 0.853 & 0.857 \\ \cline{2-10} 
 & \textbf{Rec.} & 0.317 & 0.970 & 0.913 & \textbf{0.974} & 0.883 & 0.872 & 0.901 & 0.937 \\ \cline{2-10} 
 & \textbf{$\mathbf{F}_1$} & 0.466 & 0.862 & 0.846 & 0.837 & 0.868 & 0.874 & 0.876 & \textbf{0.895} \\ \hline
\end{tabular}
\begin{tablenotes}
\item $\dagger$: Text-based methods 
\item $\wr$: Image-based methods
\item $\ddagger$: Multi-modal methods
\end{tablenotes}
\vspace{-1.5em}
\end{threeparttable}
\end{adjustbox}
\end{table}

\vspace{-1.5em}
\subsubsection{General Performance Analysis}
The general performance of $\mathsf{SAFE}$ and baselines are provided in Tab. \ref{tab:general_performance}. Results indicate when predicting fake news, $\mathsf{SAFE}$ can outperform all baselines based on the accuracy values and $F_1$ scores for both datasets. 
Based on PolitiFact data, the general performance of methods is $\mathsf{SAFE}>\text{att-RNN}\approx \text{LIWC}> \text{VGG-19}$; while for GossipCop data, such performance is $\mathsf{SAFE}>\text{VGG-19}> \text{att-RNN}>\text{LIWC}$.
Note that multiple supervised learners (such as SVM, decision tree, logistic regression, and $k$-NN) have been used with LIWC in our experiments, where we present the best performance (obtained from random forest) in Tab.~\ref{tab:general_performance}. 

% \begin{figure}[h]
%     \centering
%     \subfigure[GossipCop]{
%     \includegraphics[width=.47\textwidth]{gc_confusion.eps}} \quad
%     \subfigure[PolitiFact]{
%     \includegraphics[width=.47\textwidth]{pf_confusion.eps}}
%     \caption{General Performance based on Confusion Matrix}
%     \label{fig:confusion}
% \end{figure}

\begin{figure}[t]
\begin{minipage}{0.33\textwidth}
    \centering
    \subfigure[PolitiFact]{
    \includegraphics[width=\textwidth]{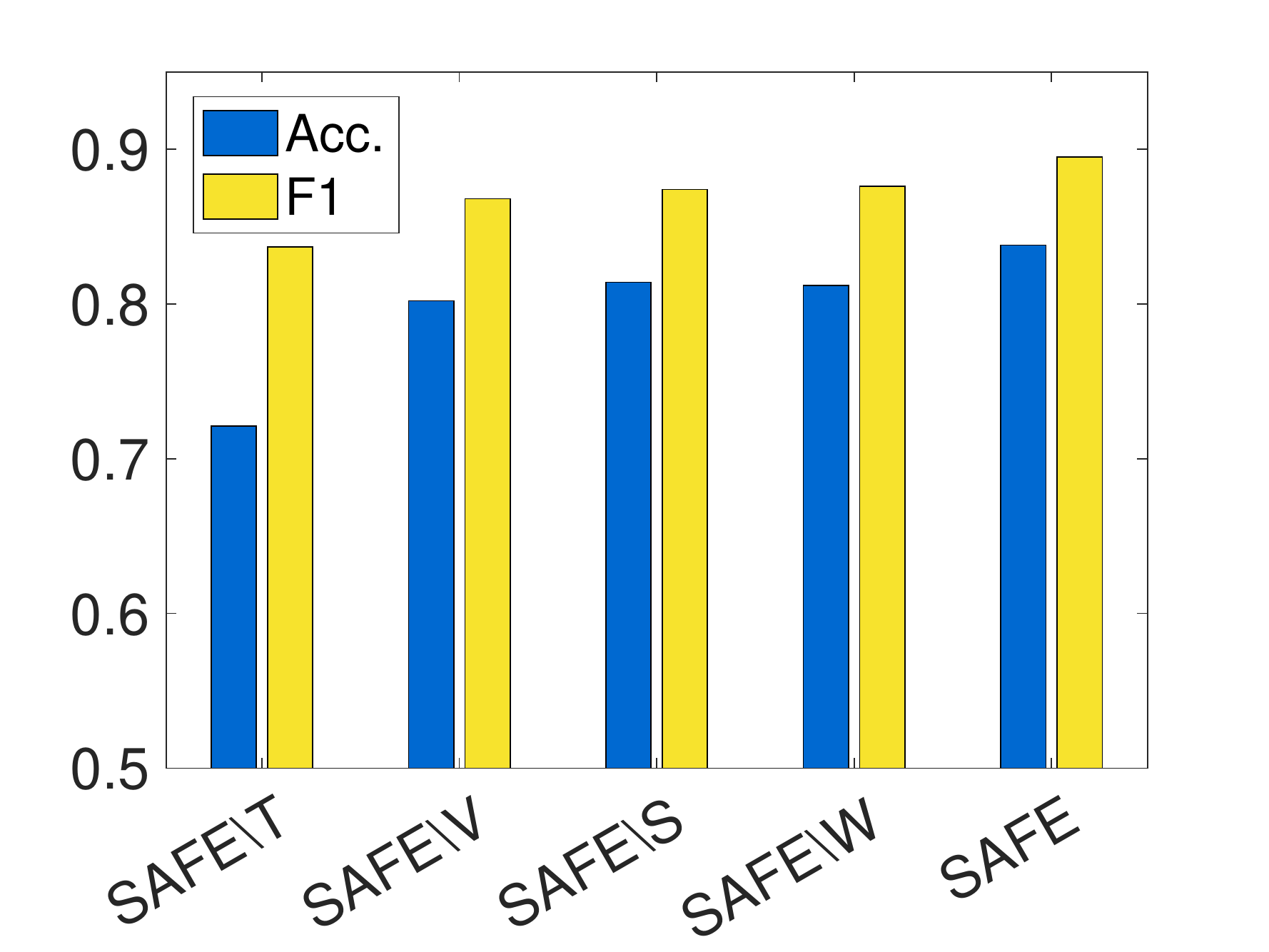}} 
    \subfigure[GossipCop]{
    \includegraphics[width=\textwidth]{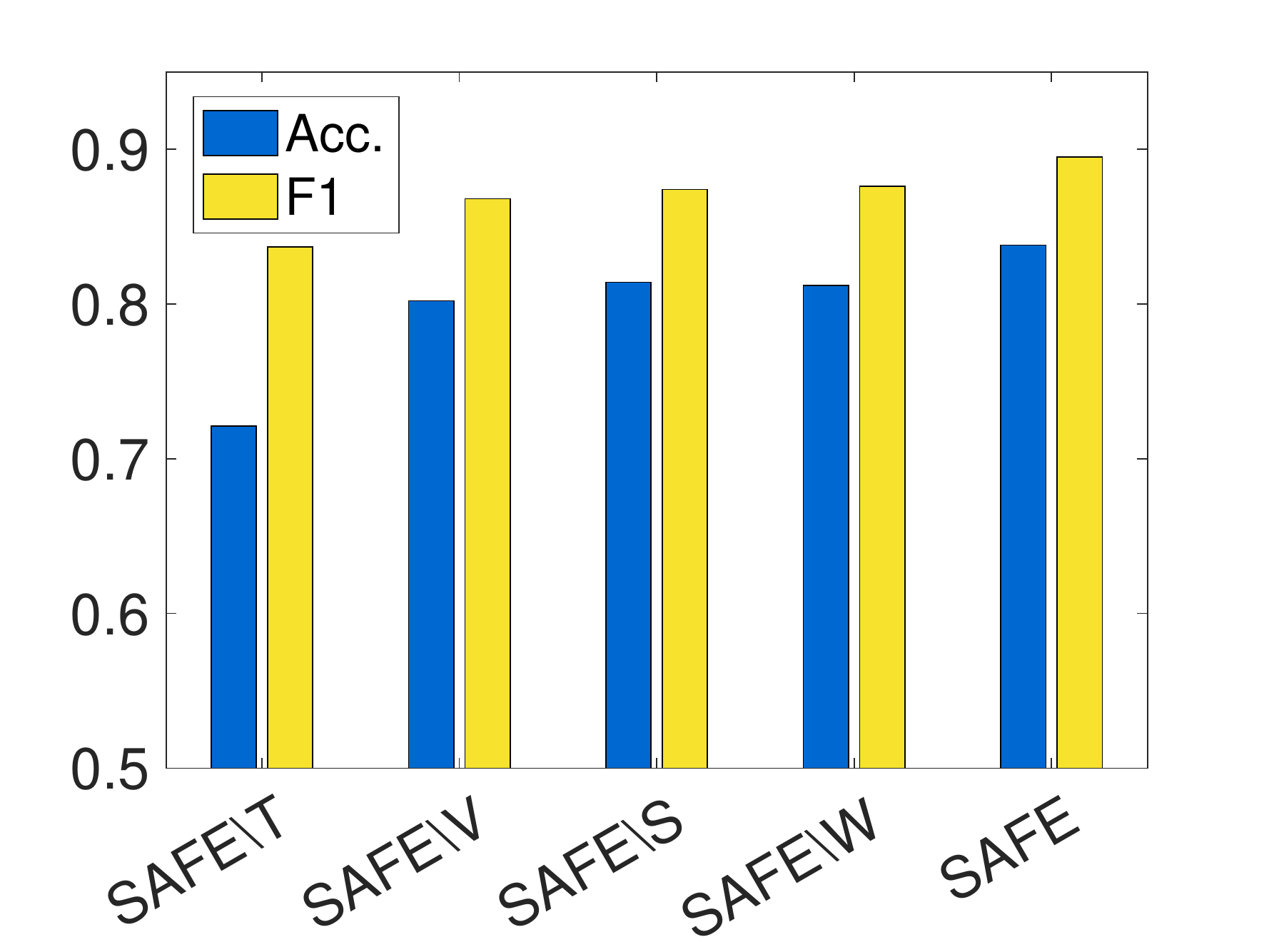}}
    \vspace{-1.5em}
    \caption{Module Analysis}
    \label{fig:module_analysis} 
\end{minipage}
\begin{minipage}{0.66\textwidth}
    \centering
    \subfigure[PolitiFact]{
    \includegraphics[width=0.5\textwidth]{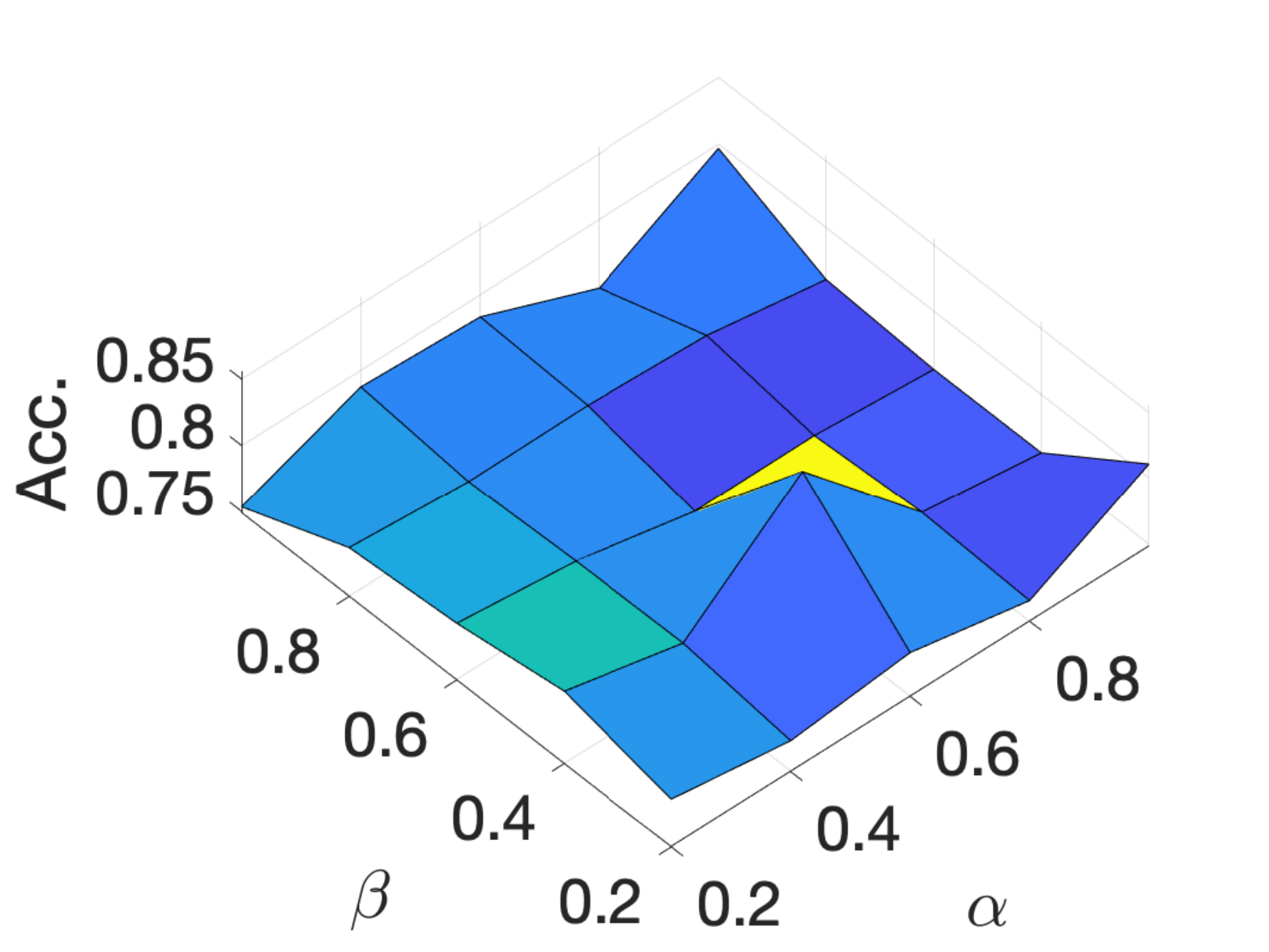}
    \includegraphics[width=0.5\textwidth]{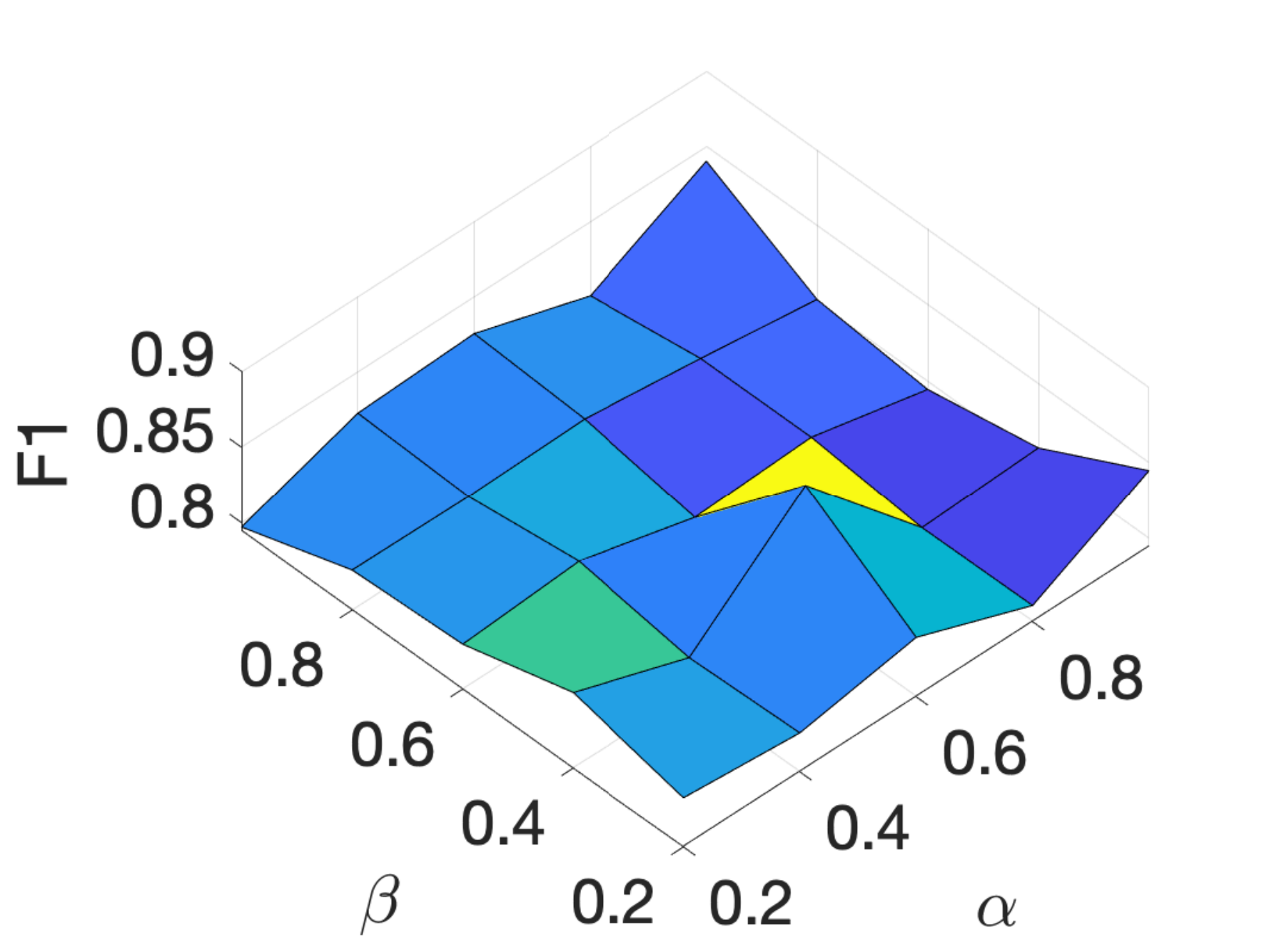}} 
    \subfigure[GossipCop]{
    \includegraphics[width=0.5\textwidth]{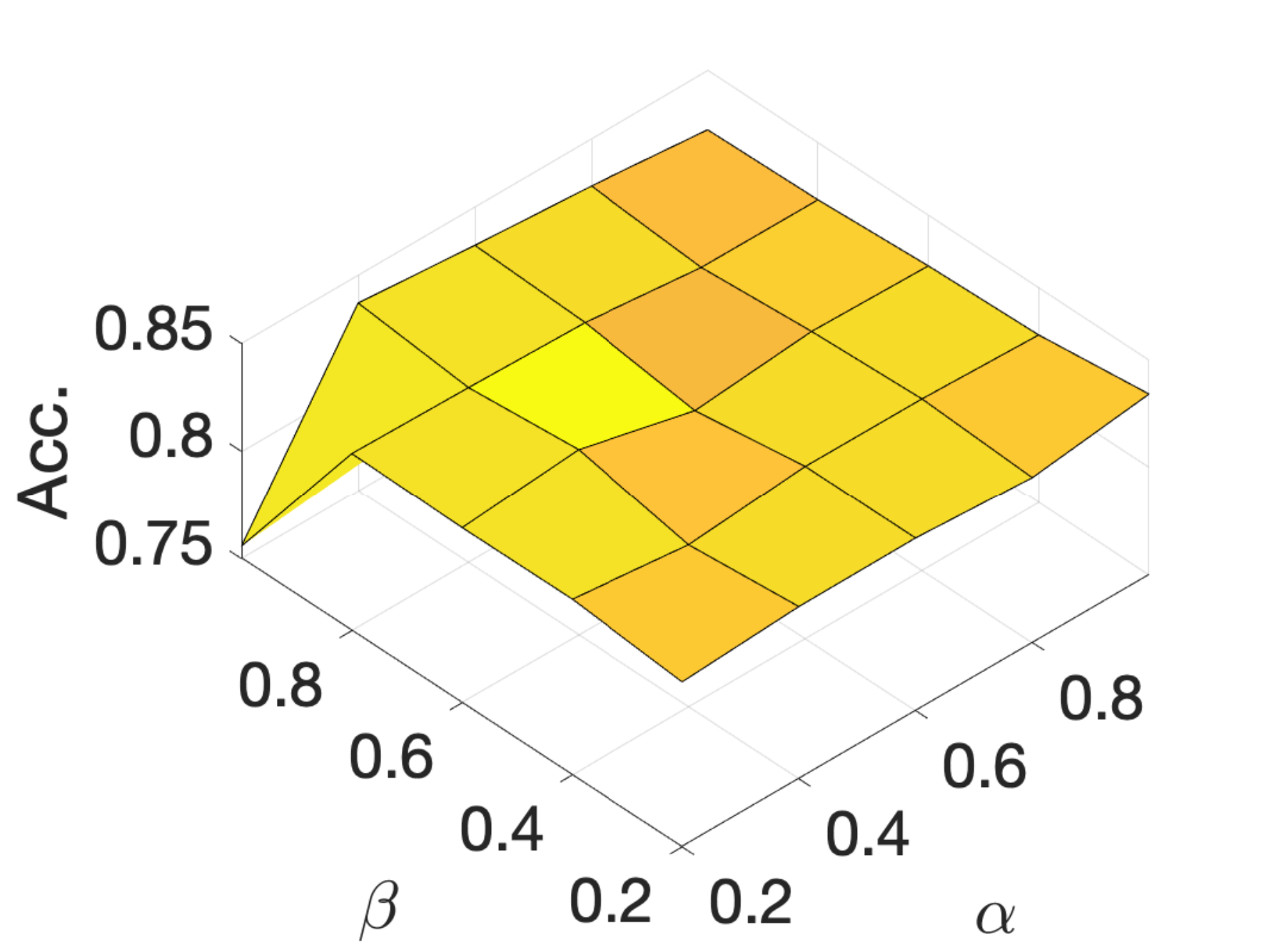}
    \includegraphics[width=0.5\textwidth]{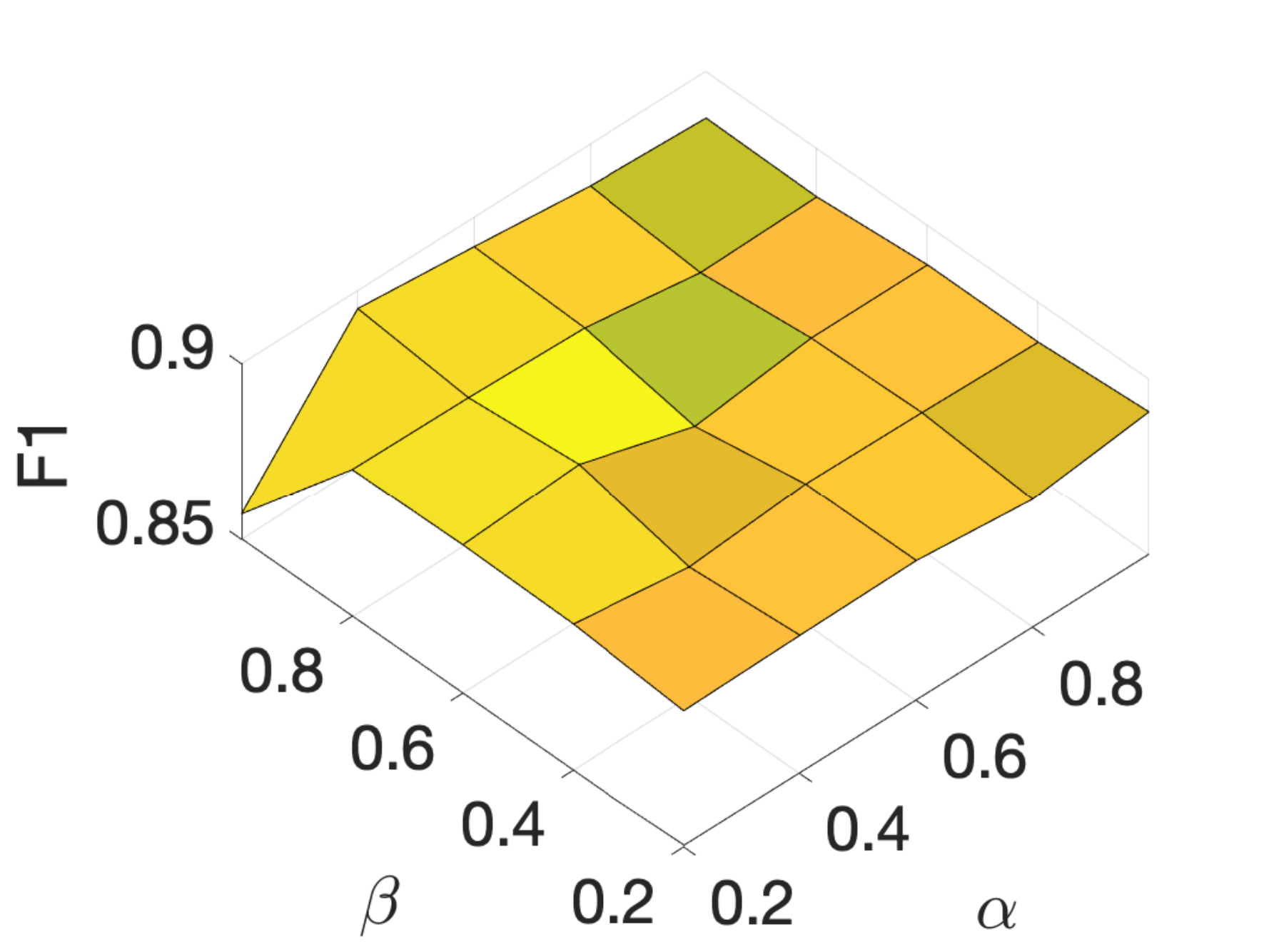}} 
    \vspace{-1.5em}
    \caption{Parameter Analysis}
    \label{fig:parameter_analysis}
\end{minipage}
\vspace{-1.5em}
\end{figure}

\vspace{-1.5em}
\subsubsection{Module Analysis} The performance of $\mathsf{SAFE}$ %($\mathsf{SAFE}$) 
and its variants 
%($\mathsf{SAFE} \setminus$T, $\mathsf{SAFE} \setminus$V, $\mathsf{SAFE} \setminus$S, and $\mathsf{SAFE} \setminus$W) 
are presented in Tab.~\ref{tab:general_performance} and Fig.~\ref{fig:module_analysis}. Results indicate when predicting fake news, (1) integrating news textual information, visual information, and their relationship ($\mathsf{SAFE}$) performs best among all variants, (2) using multi-modal information ($\mathsf{SAFE} \setminus$S or $\mathsf{SAFE} \setminus$W) performs better compared to using single-modal information ($\mathsf{SAFE} \setminus$T or $\mathsf{SAFE} \setminus$V); (3) it is comparable to detect fake news by either independently using multi-modal information ($\mathsf{SAFE} \setminus$S) or mining their relationship ($\mathsf{SAFE} \setminus$W); and (4) textual information ($\mathsf{SAFE} \setminus$V) is more important compared to visual information ($\mathsf{SAFE} \setminus$T).\vspace{-1.5em}

\subsubsection{Parameter Analysis}
In Eq. (\ref{eq:loss_function}), $\alpha$ and $\beta$ are used to allocate the relative importance between the extracted multi-modal features ($\alpha$) and the similarity across modalities ($\beta$). 
To assess their influence in method performance, we changed the value of $\alpha$ and $\beta$ respectively from 0 to 1 with a step size of 0.2. Results in Fig.~\ref{fig:parameter_analysis} show that various parameter values lead to the accuracy (or $F_1$ score) of $\mathsf{SAFE}$ ranging from 0.75 to 0.85 (or from 0.8 to 0.9) for both datasets. The proposed method performs best when $\alpha:\beta = 0.4:0.6$ in PolitiFact and $\alpha:\beta =  0.6:0.4$ in GossipCop, which again validates the importance of both multi-modal information and cross-modal relationship in predicting fake news.

\vspace{-1.5em}
\subsubsection{Case Study} 
\begin{figure}[t]
    \subfigure[$s=0.024$]{
    \label{subfig:building}
    \includegraphics[width = 0.32\textwidth]{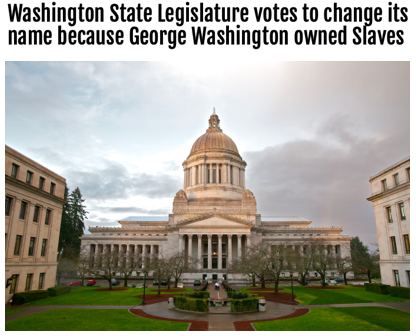}}
    \subfigure[$s=0.044$]{\label{subfig:death}
    \includegraphics[width = 0.305\textwidth]{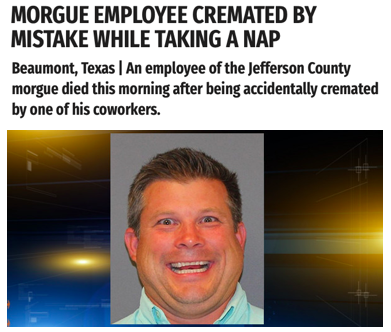}}
    \subfigure[$s=0.001$]{\label{subfig:non-couple}
    \includegraphics[width =0.335\textwidth]{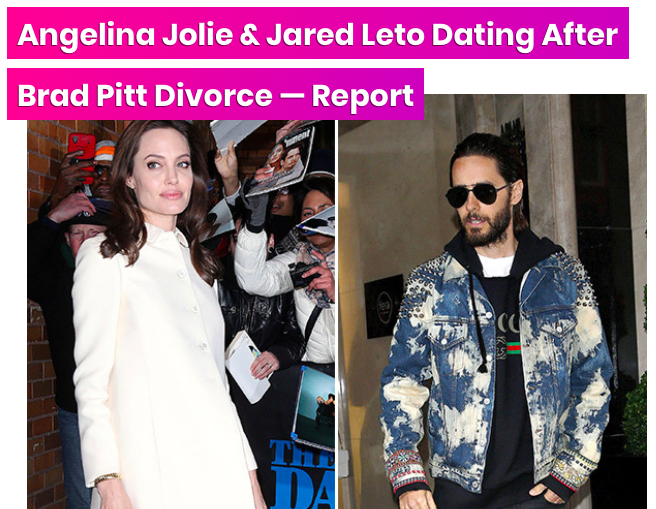}}
    \vspace{-1.5em}
    \caption{Fake News}
    \label{fig:fake_news}

    \subfigure[$s=0.966$]{
    \label{subfig:transcript}
    \includegraphics[width = 0.29\textwidth]{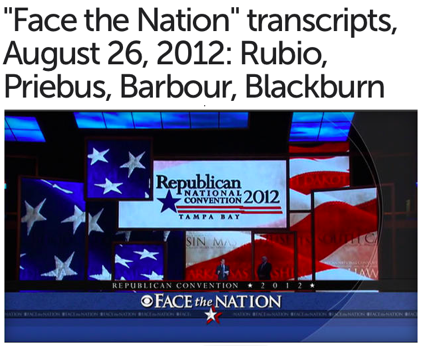}}
    \subfigure[$s=0.975$]{
    \label{subfig:parade}
    \includegraphics[width = 0.31\textwidth]{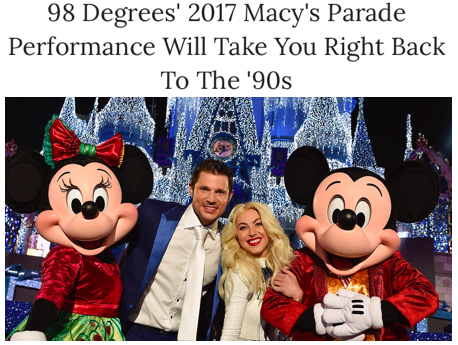}}
    \subfigure[$s=0.983$]{
    \label{subfig:couple}
    \includegraphics[width = 0.355\textwidth]{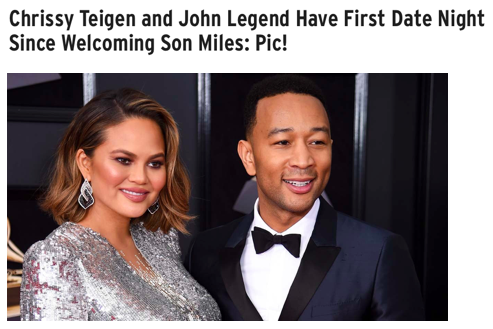}}
    \vspace{-1.5em}
    \caption{True News}
    \label{fig:true_news}
    \vspace{-1.5em}
\end{figure}

In our case studies, we aim to answer the following questions: is there any real-world fake news story whose textual and visual information are not closely related to each other? If there is, can $\mathsf{SAFE}$ correctly recognize such irrelevance and further recognize its falsity? For this purpose, we went through the news articles in the two datasets, and compared their ground truth labels with their similarity scores computed by $\mathsf{SAFE}$. Several examples are presented in Figs.~\ref{fig:fake_news}-\ref{fig:true_news}.
% with extreme computed similarity scores are presented in Figs. \ref{fig:fake_news}-\ref{fig:true_news}.
It can be observed that (\textbf{I}) the gap between textual and visual information exist for some fictitious stories for (but not limited to) two reasons. First, such stories are difficult to be supported by non-manipulated images. An example is in Fig. \ref{subfig:building}, where no voting- and bill-related image is actually available. Compared to the couples having a real intimate relationship (see Fig. \ref{subfig:couple}), the fake ones often have rare group photos or use collages (see Fig. \ref{subfig:non-couple}).
    % fake news: hard to find the real image, hence prefer to using collage -> (5-c);
    % true news: usually have a matched image (6-c)
    Second, using ``attractive'' though not closely relevant images can help increase the news traffic. For example, the fake news in Fig.~\ref{subfig:death} includes an image with a smiling individual that conflicts with the death story.
(\textbf{II}) $\mathsf{SAFE}$ helps correctly assess the relationship (similarity) between news textual and visual information. For fake news stories in Fig.~\ref{fig:fake_news}, their corresponding similarity scores are all low and $\mathsf{SAFE}$ correctly labels them as fake news. Similarly, $\mathsf{SAFE}$ assigns all true news stories in Fig. \ref{fig:true_news} a high similarity score, and predicts them as true news.
% fake news ~ small similarity
% true news ~ large similarity

% noted s in different datasets is not comparable.

\vspace{-1.5em}
\section{Conclusion}
\label{sec:conclusion}

\vspace{-1em}
In this work, a similarity-aware multi-modal method, named $\mathsf{SAFE}$,  is proposed to predict fake news. The method extracts both textual and visual features of news content, and investigates their relationship. Experimental results indicate multi-modal features and the cross-modal relationship (similarity) are valuable with a comparable importance in fake news detection. Case studies conducted further validate the effectiveness of the proposed method in assessing such similarity and predicting fake news. Nevertheless, we should point out the proposed method investigates textual and visual information without considering, e.g., network and video information. Additionally, relationships within modalities are valuable as well such as the textual (or visual) similarity among or between pairwise news articles, which both will be part of our future work.

% \begin{theorem}
% This is a sample theorem. The run-in heading is set in bold, while
% the following text appears in italics. Definitions, lemmas,
% propositions, and corollaries are styled the same way.
% \end{theorem}
%
% the environments 'definition', 'lemma', 'proposition', 'corollary',
% 'remark', and 'example' are defined in the LLNCS documentclass as well.
%
% \begin{proof}
% Proofs, examples, and remarks have the initial word in italics,
% while the following text appears in normal font.
% \end{proof}

\bibliographystyle{splncs04}

\vspace{-1em}
\bibliography{references}

\end{document}